\documentclass{article}

\usepackage{microtype}
\usepackage{graphicx}
\usepackage{subcaption}
\usepackage{booktabs} 

\usepackage{hyperref}

\usepackage[preprint]{icml2026}

\usepackage{amsmath}
\usepackage{amssymb}
\usepackage{mathtools}
\usepackage{amsthm}
\usepackage[table]{xcolor}

\usepackage{booktabs}
\usepackage{tabularx}
\usepackage{xcolor}
\usepackage{pifont}     
\usepackage{amsmath}    
\usepackage{ragged2e}   
\usepackage{enumitem}   

\usepackage[capitalize,noabbrev]{cleveref}

\theoremstyle{plain}

\theoremstyle{definition}

\theoremstyle{remark}

\usepackage{ulem}
\usepackage{xcolor}
\usepackage{multirow}

\usepackage[utf8]{inputenc} 
\usepackage[T1]{fontenc}    

\usepackage[textsize=tiny]{todonotes}
\icmltitlerunning{HELP: HyperNode Expansion and Logical Path-Guided Evidence Localization for Accurate and Efficient GraphRAG}

\definecolor{mygreen}{RGB}{0, 150, 0}
\definecolor{myred}{RGB}{200, 0, 0}

\begin{document}

\twocolumn[
  \icmltitle{HELP: HyperNode Expansion and Logical Path-Guided Evidence\\Localization for Accurate and Efficient GraphRAG}



\icmlsetsymbol{equal}{*}
\icmlsetsymbol{cor}{$\dagger$} 

\begin{icmlauthorlist}
  \icmlauthor{Yuqi Huang}{sjtu}
  \icmlauthor{Ning Liao}{sjtu}
  \icmlauthor{Kai Yang}{sjtu}
  \icmlauthor{Anning Hu}{sjtu}
  \icmlauthor{Shengchao Hu}{sjtu}
  \icmlauthor{Xiaoxing Wang}{sjtu,cor} 
  \icmlauthor{Junchi Yan}{sjtu}
\end{icmlauthorlist}

\icmlaffiliation{sjtu}{Shanghai Jiao Tong University, China}

\icmlcorrespondingauthor{Xiaoxing Wang}{figure1\_wxx@sjtu.edu.cn}


\vskip 0.3in
] 

\printAffiliationsAndNotice{$^\dagger$Corresponding author.} 

\begin{abstract}
Large Language Models (LLMs) often struggle with inherent knowledge boundaries and hallucinations, limiting their reliability in knowledge-intensive tasks. While Retrieval-Augmented Generation (RAG) mitigates these issues, it frequently overlooks structural interdependencies essential for multi-hop reasoning. Graph-based RAG approaches attempt to bridge this gap, yet they typically face trade-offs between accuracy and efficiency due to challenges such as costly graph traversals and semantic noise in LLM-generated summaries. In this paper, we propose \underline{\textbf{H}}yperNode \underline{\textbf{E}}xpansion and \underline{\textbf{L}}ogical \underline{\textbf{P}}ath-Guided Evidence Localization strategies for GraphRAG (\textbf{HELP}), a novel framework designed to balance accuracy with practical efficiency through two core strategies: 1) HyperNode Expansion, which iteratively chains knowledge triplets into coherent reasoning paths abstracted as HyperNodes to capture complex structural dependencies and ensure retrieval accuracy;
and 2) Logical Path-Guided Evidence Localization, which leverages precomputed graph-text correlations to map these paths directly to the corpus for superior efficiency. HELP avoids expensive random walks and semantic distortion, preserving knowledge integrity while drastically reducing retrieval latency. Extensive experiments demonstrate that HELP achieves competitive performance across multiple simple and multi-hop QA benchmarks and up to a 28.8$\times$ speedup over leading Graph-based RAG baselines.

\end{abstract}

\section{Introduction}
Large Language Models (LLMs) have demonstrated strong capabilities in natural language understanding and generation, but they are prone to hallucination and often lack up-to-date domain knowledge, which limits their reliability in knowledge-intensive applications~\cite{huang2025survey}. A common solution is Retrieval-Augmented Generation (RAG), which splits documents into passages, retrieves the top‑k relevant passages via dense vector search, and appends them to the prompt to ground the LLM’s response~\cite{gao2023retrieval, lewis2020retrieval}. This pipeline is simple, robust, and scalable, making it practical for real-world deployment. However, dense retrieval often treats seemingly unstructured text as flat content and overlooks the underlying structure that can be distilled from it~\cite{han2024retrieval}. This structured information is crucial for organizing and retrieving knowledge in massive, multi-source corpora, which limits the effectiveness of dense retrieval in realistic settings~\cite{peng2024graph}. Many queries require multi-step reasoning~\cite{plaat2025multi} and a deep understanding of the interconnections between entities and events across disparate sources. In these cases, RAG’s similarity-based top‑k retrieval emphasizes local paragraph relevance rather than knowledge structure, which can introduce redundant or unfocused context and reduce performance on multi-hop and relational questions~\cite{zhang2025survey}.

To bridge this structural gap, Graph-based RAG approaches have been proposed to incorporate Knowledge Graphs into the retrieval process. The standard workflow of Graph-based RAG~\cite{gao2023retrieval,zhang2025survey} typically involves two stages: Knowledge Graph Construction and Graph-Augmented Retrieval. In the construction phase, entities and their corresponding relations are extracted from raw corpus to build a structured graph~\cite{edge2024local, jimenez2024hipporag, gutierrez2025rag}, though some lightweight variants~\cite{zhuang2025linearrag} focus exclusively on entity indexing. During retrieval, these methods navigate the relational topology using graph algorithms such as Personalized PageRank (PPR)~\cite{yang2024efficient} to identify relevant evidence, which are then synthesized into textual context for the LLM.

However, existing solutions face a dilemma between accuracy and efficiency~\cite{han2025rag, xiang2025use}. \textbf{(1) Graph-based RAG methods can suffer from inefficient retrieval of information stored in knowledge graphs.} Methods such as HippoRAG~\cite{jimenez2024hipporag}, HippoRAG2~\cite{gutierrez2025rag} and ToG~\cite{sun2023think, ma2024think} preserve rich relational information for complex reasoning, but their reliance on graph traversal introduces substantial computational overhead and latency. Moreover, GraphRAG’s LLM-generated community summaries~\cite{edge2024local} can introduce semantic noise and secondary hallucinations, degrading retrieval precision. 
\textbf{(2) Some methods ignores edge information to simplify retrieval and improve efficiency, but at the cost of accuracy.} 
LinearRAG~\cite{zhuang2025linearrag} accelerates retrieval by constructing a relation-free hierarchical graph, termed Tri-Graph, using only lightweight entity extraction and semantic linking, effectively bypassing complex relational modeling. Similarly, $E^2$GraphRAG~\cite{zhao20252graphrag} streamlines the indexing process by utilizing SpaCy to construct bidirectional indexes between entities and passages to capture their many-to-many relations. While these approaches achieve significant speedups in indexing and retrieval, their reliance on simplified topologies weakens the explicit semantic dependencies required for precise multi-hop reasoning. Thus, a solution is still needed that maintains high accuracy while improving efficiency.


We propose \underline{\textbf{H}}yperNode \underline{\textbf{E}}xpansion and \underline{\textbf{L}}ogical \underline{\textbf{P}}ath-Guided Evidence Localization strategies for GraphRAG, named as \textbf{HELP}. It is a relation-aware retrieval framework that balances multi-hop accuracy with practical efficiency. On the one hand, we introduce HyperNode as a core unit to represent and chain knowledge triplets for multi-hop reasoning. In this framework, each retrieved knowledge triplet is instantiated as a HyperNode, which is then used to update the working graph, enabling iterative retrieval that chains HyperNodes together. This process naturally expands reasoning paths, transforming isolated facts into multi-hop relational chains.
On the other hand, we introduce Logical Path-Guided Evidence Localization, an efficient strategy that utilizes the expanded final HyperNodes as precise evidence to fetch the most relevant supporting passages from the corpus. By first constructing the relational path in the structured space, this approach converts the complex multi-hop question into a targeted retrieval task, ensuring that the returned passages are both semantically relevant and logically connected.
Extensive experiments show that our method can preserve the structural integrity of knowledge graphs while maintaining the high retrieval efficiency. Our contributions are summarized as follows:


\textbf{1) An Iterative HyperNode Retrieval Strategy for Reasoning Path Expansion.} We introduce the HyperNode, a higher-order retrieval unit that bundles triples together with their relational paths into a unified entity. By explicitly modeling knowledge paths rather than isolated fragments, HyperNode better captures inter-fact dependencies and substantially improves multi-hop reasoning, while maintaining strong performance on single-hop QA.
    

\textbf{2) An Efficient Reasoning Path-Guided Evidence Localization Strategy.} 
We propose a lightweight Path-Guided Evidence Localization strategy that leverages precomputed graph-text correlations. 
By mapping the HyperNodes constructed in the structured space directly to the source corpus, this strategy converts complex multi-hop questions into targeted retrieval tasks. 



\textbf{3) Superior Performance and Consistent Effectiveness.} Our framework achieves an excellent balance between multi-hop reasoning accuracy and computational efficiency. Crucially, our method demonstrates strong and consistent effectiveness across a variety of benchmarks under the same configuration. This consistent effectiveness renders our method a compelling, low-overhead enhancement across a wide array of Large Language Models.

\section{Related Work}

\textbf{RAG for Complex Reasoning.}
Retrieval-Augmented Generation (RAG) serves as a robust paradigm to mitigate hallucinations by grounding Large Language Models (LLMs) in external knowledge corpora~\cite{lewis2020retrieval}. While standard dense retrieval frameworks excel at addressing simple factoid queries, they often falter in complex, multi-hop reasoning scenarios where evidence is fragmented across multiple documents~\cite{yang2018hotpotqa, thakur2021beir}. To overcome these limitations, iterative retrieval methods such as IR-CoT~\cite{trivedi2023interleaving} have been proposed to interleave reasoning traces with retrieval steps. However, these approaches primarily operate on a flat semantic space, which lack explicit modeling of the structural dependencies between entities. This limitation underscores the necessity of structured retrieval mechanisms, such as Graph-based RAG~\cite{edge2024local}, which leverage relational priors to maintain logical coherence across disparate information nodes.

\textbf{Graph-based RAG.}
Recognizing the limitations of standard RAG, recent works incorporate Knowledge Graphs (KGs) to introduce structural priors~\cite{hu2025graggraphretrievalaugmentedgeneration, mavromatis2024gnnraggraphneuralretrieval}.
Approaches like HippoRAG~\cite{jimenez2024hipporag} and its successor HippoRAG2~\cite{gutierrez2025rag} draw inspiration from hippocampal memory indexing, utilizing Personalized PageRank algorithm to navigate relational paths. By mimicking associative memory, they effectively bridge disconnected entities across the corpus. At the same time, methods such as GraphRAG~\cite{edge2024local} focus on hierarchical abstraction; they employ LLMs to pre-summarize clustered graph communities, providing a "map-reduce" style global context that handles high-level thematic queries more effectively than point-wise retrieval.
In parallel, GNN-RAG~\cite{mavromatis2024gnnraggraphneuralretrieval} introduces a neuro-symbolic framework that utilizes GNNs to reason over dense subgraphs for answer retrieval, verbalizing the extracted shortest paths into natural language to enable LLMs to perform complex multi-hop reasoning.
Despite their effectiveness in improving reasoning accuracy, these methods typically incur high computational costs and latency due to complex graph algorithms or extensive LLM usage, limiting their scalability in real-time applications.


\textbf{Efficient Retrieval with Structural Awareness.}
Balancing structural utilization with retrieval efficiency remains a formidable challenge. Current research primarily diverges into two extremes. On one hand, lightweight frameworks like LinearRAG~\cite{zhuang2025linearrag} achieve rapid retrieval by simplifying graph topology; however, this reductionist approach inevitably discards critical relational semantics and high-order dependencies necessary for precise multi-hop reasoning. On the other hand, structure-centric approaches prioritize reasoning depth but frequently encounter scalability bottlenecks. For instance, path-based methods such as ToG~\cite{sun2023think} rely on iterative LLM-guided beam searches, which capture complex dependencies at the cost of prohibitive inference latency. Similarly, subgraph-based methods like QA-GNN~\cite{yasunaga2021qa} and GRAG~\cite{hu2025graggraphretrievalaugmentedgeneration} try to preserve relational contexts via $k$-hop ego networks, yet incur substantial overhead during subgraph extraction and encoding. To bridge this gap, hybrid approaches~\cite{jin2024graph, jiang2025kg} use adaptive agents to dynamically select retrieval granularities (nodes, paths, or subgraphs). While these methods mitigate noise, the decision-making process of agents often introduces extensive overhead that limits real-time applicability. 

\section{Methodology}\label{sec:methodology}

\begin{figure*}[!t] 
    \centering
    \vspace{-10pt}
    \includegraphics[width=0.95\textwidth]{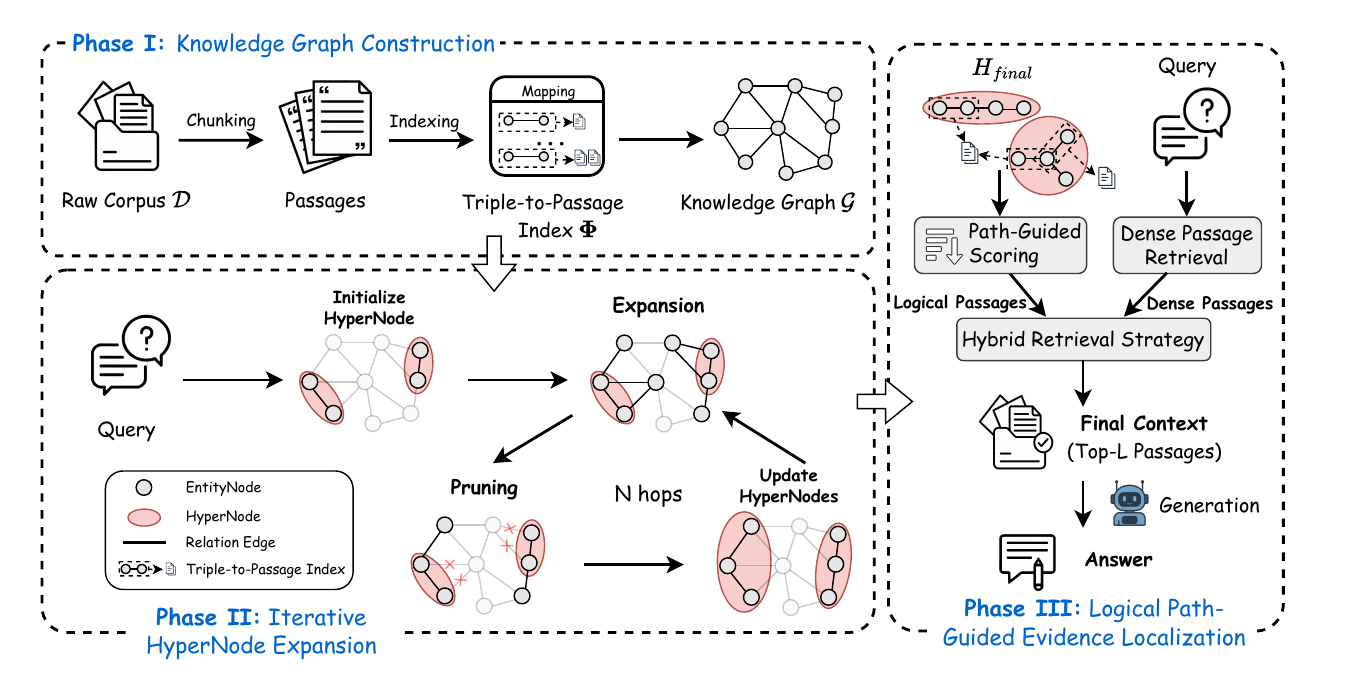} 
    \vspace{-8pt}
    \caption{Overview of the HELP framework. The workflow consists of three stages: (I) Knowledge Graph Construction, utilizing OpenIE to build the Triple-to-Passage Index; (II) Iterative HyperNode Expansion, which iteratively chains triples as HyperNodes while pruning irrelevant ones to maintain reasoning coherence; and (III) Logical Path-Guided Evidence Localization, where HyperNodes are grounded back to the original passages for precise evidence retrieval via a hybrid mechanism.}
    \label{fig: HELP framework}
\vspace{-10pt}
\end{figure*}

We elaborate on the mechanism of HELP, as illustrated in Fig.~\ref{fig: HELP framework}. Our approach comprises of three phases: Knowledge Graph Construction (Sec.~\ref{sec: knowledge_graph_construction}), Iterative HyperNode Expansion (Sec.~\ref{sec:hypernode_expansion}), and Logical Path-Guided Evidence Localization (Sec.~\ref{sec:evidence_localization}).

\subsection{Knowledge Graph Construction}\label{sec: knowledge_graph_construction}
To incorporate structural knowledge, Graph-Based RAG leverages a Knowledge Graph denoted as $\mathcal{G}=(\mathcal{V}, \mathcal{E})$, where $\mathcal{V}$ represents entities and $\mathcal{E}$ represents relations. The graph is typically constructed by extracting triplets from the corpus. Existing methods perform graph traversal, such as Personal PageRank or BFS, starting from entities mentioned in $q$. However, traversing the entire graph topology is computationally expensive, and ignoring edge semantics leads to accuracy loss. Our method aims to resolve this trade-off.

To bridge the gap between unstructured text and structured reasoning, we first construct a Knowledge Graph $\mathcal{G}$. Given a corpus of documents, we partition each document into passages $\mathcal{P} = \{p_1, p_2, \dots, p_N\}$. Then we employ an Open Information Extraction (OpenIE) module to extract a set of relational triplets from each passage $p_i$. Let $\mathcal{T}_i = \{ (h_k, r_k, t_k) \}_{k=1}^{|\mathcal{T}_i|}$ denote the sequence of relational triplets extracted from passage $p_i$.

Crucially, to support our retrieval mechanism, we construct a Triple-to-Passage Index $\Phi$, which captures graph-text correlations. For every unique triplet $\tau = (h, r, t)$, we maintain a mapping to its provenance passages:
\begin{equation}
    \Phi(\tau) = \left\{ (p_i, w_i) \mid \tau \in \mathcal{T}_i, \quad w_i = \frac{1}{|\mathcal{T}_i|} \right\}
\end{equation}
$w_i$ serves as a density-normalized weight, ensuring passages containing fewer yet more specific facts outweigh dense, noisy passages in retrieval scores.

\subsection{Iterative HyperNode Expansion}
\label{sec:hypernode_expansion}

To enable multi-hop reasoning, we introduce HyperNode Expansion, a strategy built upon the iterative merging and expanding of relational contexts. In our framework, a HyperNode $H = \{\tau_1, \dots, \tau_m\}$ is defined as a cumulative unit instantiated by merging $m$ coherent knowledge triplets into a unified semantic representation. By recursively expanding these nodes with adjacent relations, we transform discrete, isolated facts into integrated multi-hop reasoning paths.

\subsubsection{HyperNode Serialization}

To map a HyperNode into a dense vector space, we must handle the permutation invariance of the set structure. Unlike traditional sequential data, a HyperNode $H$ is an unordered set of triplets. Without a canonical ordering, the encoder $E(\cdot)$ might produce inconsistent embeddings for the same set of facts  in different sequences, hindering the effective representation of evidence within the structured data.

To mitigate this, we employ a deterministic linearization function $S(\cdot)$ designed to bridge the gap between structured sets and sequential modeling. Specifically, we first sort the triplets lexicographically based on their constituent elements to ensure a unique, reproducible order regardless of the original data extraction sequence. Once sorted, these structured components are flattened into a unified text sequence:
\begin{equation}
    S(H) = \text{join}\left( \bigcup_{\tau \in \text{sorted}(H)} \text{concat}(h_\tau, r_\tau, t_\tau) \right)
\end{equation}
The dense vector representation of the HyperNode is then obtained via a pre-trained Transformer-based encoder: $v_{H} = E(S(H))$. By mapping the relational structure into a continuous latent space, this process allows the model to capture the aggregate semantic context. 

\subsubsection{Iterative Expansion Process}
The core of our approach is the iterative expansion mechanism. Given a user query $q$, we first map it to a normalized vector representation $v_q = \frac{E(q)}{\|E(q)\|_2}$, where $E(\cdot)$ denotes the encoder. The expansion process then updates the set of active HyperNodes over $N$ hops to generate the final set $H_{final}$. The overall procedure is summarized in Algorithm \ref{alg:hypernode_expansion} and details are provided below.

\textbf{HyperNode Initialization.} It begins with the identification of seed HyperNodes, which serve as the semantic anchors for subsequent reasoning. Instead of starting from raw entities, we leverage the fine-grained semantics of individual triplets $\{\tau\}$ extracted from the corpus. We compute the cosine similarity between the query vector $v_q$ and the normalized embedding of each candidate triplet. By selecting the top-$n$ triplets with the highest alignment scores, we form the initial set $H_{1}$. This step ensures that the expansion starts from the most promising factual kernels, effectively reducing the noise introduced by irrelevant graph regions.


\textbf{Iterative Expansion.} At each step $i$ (where $2 \le i \le N$, and $N$ denotes the number of expansion hops), the model extends reasoning paths to incorporate broader context. Specifically, in the $i$-th expansion round, we process each HyperNode $H \in H_{i-1}$ individually. We identify a set of adjacent triplets $\mathcal{T}_{adj}$ from the Knowledge Graph $\mathcal{G}$ that are directly connected to the entities contained within the current $H$. For each adjacent triplet $\tau_{next} \in \mathcal{T}_{adj}$, we instantiate a new expanded HyperNode $H' = H \cup \{\tau_{next}\}$. The global candidate set $\mathcal{C}_{cand}$ is then formed by accumulating all such generated $H'$ instances derived from each preceding HyperNode in $H_{i-1}$. This incremental growth allows each HyperNode to transition from a single fact to a complex path structure representing a multi-hop chain of evidence.


\textbf{Pruning via Semantic Distance.} To counteract the exponential growth of the search space and to maintain focus on the query's core intent, we implement a semantic-guided beam search strategy. This strategy ensures that only the most semantically pertinent reasoning paths are retained at each hop. We serialize the content of each candidate $H'$ into a sequence $S(H')$ and compute its distance to the query. Since the embeddings are normalized, we utilize the Euclidean distance as our metric:
\begin{equation}
    \label{eq:dist_metric}
    \text{dist}(H', q) = \left\| \frac{E(S(H'))}{\|E(S(H'))\|_2} - v_q \right\|_2
\end{equation}
Based on Eq. \eqref{eq:dist_metric}, we select the top-$k$ HyperNodes with the minimal distance to form $H_{i}$. By maintaining a fixed beam width $k$, it constrains the search space to a manageable scale, preventing the computational overhead typically associated with deep graph traversals. Consequently, it ensures that the final converged set, $H_{final}$, is not merely a collection of isolated facts, but a structured assembly of complete, logically coherent, and semantically aligned reasoning paths tailored to the specific information need.

\begin{algorithm}[!tb]
\caption{Iterative HyperNode Expansion Algorithm}
\label{alg:hypernode_expansion}
\begin{algorithmic}[1]
\REQUIRE User query $q$, Knowledge Graph $\mathcal{G}$, Expansion hops $N$, Initial triple size $n$, Beam size $k$
\ENSURE Final set of HyperNodes $H_{final}$

\STATE \textbf{Initialize:} $v_q \leftarrow \text{Normalize}(E(q))$

\STATE \textit{// Step 1: Initialization with Seed HyperNodes}
\STATE Retrieve all triples $\{\tau\}$ from $\mathcal{G}$
\STATE Score triples by similarity to $v_q$
\STATE $H_1 \leftarrow \text{Top-}n \text{ ranked triples}$

\STATE \textit{// Step 2: Iterative Expansion Process}
\FOR{$i = 2$ to $N$}
    \STATE $\mathcal{C}_{cand} \leftarrow \emptyset$ 
    
    \FOR{each HyperNode $H$ in $H_{i-1}$}
        \STATE $\mathcal{T}_{adj} \leftarrow \{ \tau \in \mathcal{G} \mid \tau \text{ is adjacent to entities in } H \}$
        \FOR{each $\tau_{next}$ in $\mathcal{T}_{adj}$}
            \STATE \textbf{Expand:} $H' \leftarrow H \cup \{ \tau_{next} \}$
            \STATE Add $H'$ to $\mathcal{C}_{cand}$
        \ENDFOR
    \ENDFOR
    
    \STATE \textit{// Step 3: Pruning (Beam Search)}
    \STATE $H_{i} \leftarrow \emptyset$
    \FOR{each candidate $H'$ in $\mathcal{C}_{cand}$}
        \STATE  $v_{H'} \leftarrow \text{Normalize}(E(S(H')))$
        \STATE \textbf{Score:} Calculate $\text{dist}(H', q)$ using Eq. \eqref{eq:dist_metric}
    \ENDFOR
    
    \STATE $H_{i} \leftarrow$ Select top-$k$ HyperNodes from $\mathcal{C}_{cand}$ with minimal distance
\ENDFOR

\STATE \textbf{return} $H_{final} = H_{N}$
\end{algorithmic}
\vspace{-2pt}
\end{algorithm}

\subsection{Logical Path-Guided Evidence Localization}
\label{sec:evidence_localization}

The final phase of our framework is Logical Path-Guided Evidence Localization, a process designed to transform abstract structural information into concrete textual evidence. By utilizing the expanded HyperNodes as precise semantic anchors, we can pinpoint the most relevant supporting passages within the massive corpus, ensuring that the downstream generation is grounded in verified facts.

To bridge the semantic gap between the retrieved structural knowledge and the original source corpus, we leverage the Triple-to-Passage Index $\Phi$. This specialized inverted index, as constructed in Sec.~\ref{sec: knowledge_graph_construction}, establishes a one-to-many mapping between each unique triplet $\tau$ and the set of source passages from which it was extracted. By recognizing that a single factual claim can be instantiated across multiple disparate passages, this index ensures comprehensive evidence coverage. Such a mechanism enables robust provenance tracking, allowing the model to backtrack from a specific hop in the reasoning path to all supporting textual fragments that validate the underlying logic.

We compute a relevance score for each passage $p$ by aggregating the evidentiary signals from the complete set of reasoning paths. Instead of considering only unique triplets, we sum the contributions of all triplets $\tau$ across the final set of HyperNodes $H_{final}$ to account for the consensus of evidence. The relevance score for passage $p$ is derived as:
\vspace{-2pt}
\begin{equation}
    \small
    \text{Score}(p) = \sum_{H \in H_{final}} \sum_{\tau \in H} \mathbb{I}(p \in \Phi(\tau)) \cdot e^{-\text{dist}(H, q)} \cdot w_{p, \tau}
\end{equation}
where $w_{p,\tau}$ represents the provenance weight associated with the extraction density, as mentioned in Sec.~\ref{sec: knowledge_graph_construction}.

In this formulation, the term $\kappa(H, q) = e^{-\text{dist}(H, q)}$ functions as a non-linear soft-matching mechanism. 
By aggregating scores across all HyperNodes, this approach naturally yields a consensus-based ranking: passages are scored higher not only for semantic alignment with the query, but also for being repeatedly reinforced through multiple distinct reasoning paths.
This ensures that the evidence localization is robust to individual triplet noise and favors passages that reside at the intersection of diverse logical chains.

Drawing inspiration from hybrid search paradigms \cite{bhagdev2008hybrid, wang2025balancing}, we adopt a composite strategy to construct a robust final context of size $K$. We recognize that Logical Path-Guided Evidence Localization and Dense Passage Retrieval (DPR) operate on distinct and complementary relevance signals. Our logical path-guided approach ensures high precision by leveraging the consensus of multi-hop reasoning chains: passages that are reinforced by multiple intersecting paths receive prioritized ranking through our additive scoring mechanism. In contrast, DPR excels at capturing broad semantic coverage that structural methods might overlook.

To leverage this complementarity, we assign a primary quota of $M$ passages to the evidence localized via the Logical Path-Guided Evidence Localization, ensuring the final context is anchored in structured reasoning. This foundation is then supplemented with top-ranked candidates from DPR to fill the remaining slots, with a deduplication step to maximize incremental information gain. As demonstrated in our ablation study (see Sec.~\ref{sec: hybrid strategy}), this hybrid retrieval strategy—prioritizing structure-based consensus while backfilling with semantic matches—yields superior performance compared to single-stream methods by balancing the depth of logical rigor with the breadth of semantic recall.

\section{Experiments}

\begin{table*}[t!]
    \centering
    \caption{Performance (F1 scores) comparison of various retrieval methods across different tasks, with all models evaluated using \textit{Llama-3.3-70B-Instruct} as the backbone. Results are directly reported from HippoRAG2 \cite{gutierrez2025rag} by default, while methods marked with ($^*$) denote our local reproduction. The best results are in \textbf{bold}, and the second best are \underline{underlined}. The $^\dagger$ symbol indicates a failure in graph construction due to large corpus size.}    \label{tab:results}
    \small
    \vspace{-5pt}
    \begin{tabular}{l cccccc c}
        \toprule
        & \multicolumn{2}{c}{Simple QA} & \multicolumn{4}{c}{Multi-Hop QA} & \\
        \cmidrule(lr){2-3} \cmidrule(lr){4-7}
        Retrieval & NQ & PopQA & MuSiQue & 2Wiki & HotpotQA & LV-Eval & Avg \\
        \midrule
        \multicolumn{8}{c}{\textit{Simple Baselines}} \\
        \midrule
        None & 54.9 & 32.5 & 26.1 & 42.8 & 47.3 & 6.0 & 34.9 \\
        Contriever \cite{izacard2021unsupervised} & 58.9 & 53.1 & 31.3 & 41.9 & 62.3 & 8.1 & 42.6 \\
        BM25 \cite{robertson1994some} & 59.0 & 49.9 & 28.8 & 51.2 & 63.4 & 5.9 & 43.0 \\
        GTR \cite{ni2022large} & 59.9 & 56.2 & 34.6 & 52.8 & 62.8 & 7.1 & 45.6 \\
        \midrule
        \multicolumn{8}{c}{\textit{Large Embedding Models}} \\
        \midrule
        GTE-Qwen2-7B \cite{ni2022large} & 62.0 & 56.3 & 40.9 & 60.0 & 71.0 & 7.1 & 49.6 \\
        GritLM-7B \cite{muennighoff2024generative} & 61.3 & 55.8 & 44.8 & 60.6 & 73.3 & 9.8 & 50.9 \\
        NV-Embed-v2 \cite{lee2024nv} & 61.9 & 55.7 & 45.7 & 61.5 & 75.3 & 9.8 & 51.7 \\
        \midrule
        \multicolumn{8}{c}{\textit{Graph-based RAG Methods}} \\
        \midrule
        RAPTOR \cite{sarthi2024raptor} & 50.7 & 56.2 & 28.9 & 52.1 & 69.5 & 5.0 & 43.7  \\
        GraphRAG \cite{edge2024local} & 46.9 & 48.1 & 38.5 & 58.6 & 68.6 & 11.2 & 45.3 \\
        LightRAG \cite{guo2024lightrag} & 16.6 & 2.4 & 1.6 & 11.6 & 2.4 & 1.0 & 5.9 \\
        HippoRAG \cite{jimenez2024hipporag} & 55.3 & 55.9 & 35.1 & 71.8 & 63.5 & 8.4 & 48.3 \\
        HippoRAG2 \cite{gutierrez2025rag} & \underline{63.3} & \underline{56.2} & \textbf{48.6} & \underline{71.0} & \underline{75.5} & \textbf{12.9} & \underline{54.6} \\
        LinearRAG$^*$ \cite{zhuang2025linearrag} & 54.5 & 26.0 & 31.0 & 54.9 & 59.0 & 10.3 & 39.3 \\
        HyperGraphRAG$^*$ \cite{luo2025hypergraphrag} & 53.6 & 26.4 & 41.6 & 61.2 & 73.1 & N/A$^\dagger$ & - \\
        \midrule
        \rowcolor{blue!10} \textbf{HELP (Ours)} & \textbf{63.5} & \textbf{57.6} & \underline{48.4} & \textbf{73.9} & \textbf{75.6} & \underline{12.5} & \textbf{55.3} \\
        \bottomrule
    \end{tabular}
\vspace{-5pt}
\end{table*}

We conduct comprehensive experiments to verify the accuracy and efficiency of HELP across a diverse range of single-hop and multi-hop QA benchmarks. By comparing HELP against traditional dense retrievers, embedding models, and current Graph-based RAG frameworks, we aim to demonstrate its superior ability to localize high-quality evidence. Furthermore, we provide in-depth ablation studies to validate our Hypernode Expansion and Logical Path-Guided Evidence Localization Strategies.
\subsection{Experimental Settings}

\textbf{Datasets.} 
Our evaluation spans both Simple QA and Multi-Hop QA tasks to ensure a comprehensive assessment. For Simple QA, we utilize NaturalQuestions (NQ) dataset \cite{wang2024rear} and PopQA \cite{mallen-etal-2023-trust}; for the more challenging Multi-Hop QA, we include MuSiQue \cite{trivedi-etal-2022-musique}, 2WikiMultiHopQA(2Wiki) \cite{ho2020constructing}, HotpotQA \cite{yang2018hotpotqa}, and LV-Eval \cite{yuan2024lv}. By adopting identical data splits, corpora, and evaluation protocols as HippoRAG2~\cite{gutierrez2025rag}, we ensure a rigorous and fair comparison across both retrieval quality and generation accuracy.

\textbf{Baselines.} We compare HELP against a comprehensive set of baselines categorized into three groups, following the experimental protocol established by HippoRAG2 \cite{gutierrez2025rag}. First, we include classic and dense retrieval methods, namely BM25 \cite{robertson1994some}, Contriever \cite{izacard2021unsupervised}, and GTR \cite{ni2022large}. Second, we evaluate against state-of-the-art 7B-parameter embedding models that lead the BEIR benchmark\cite{thakur2021beir}, including GTE-Qwen2-7B-Instruct \cite{li2023towards}, GritLM-7B \cite{muennighoff2024generative}, and NV-Embed-v2 \cite{lee2024nv}. Finally, we compare HELP with Graph-based RAG methods, encompassing hierarchical approaches like RAPTOR \cite{sarthi2024raptor}, global summarization methods such as GraphRAG \cite{edge2024local} and LightRAG \cite{guo2024lightrag}, and knowledge-graph-based frameworks like HippoRAG \cite{jimenez2024hipporag}. For these aforementioned baselines, we directly report the results published in HippoRAG2 as they share an identical evaluation setting with our work. Furthermore, to ensure a timely and rigorous comparison, we faithfully reproduce the most recent advancements, including LinearRAG \cite{zhuang2025linearrag}, and HyperGraphRAG \cite{luo2025hypergraphrag}, under the same configurations to maintain complete experimental parity.

\textbf{Evaluation Metrics.} 
Consistent with the evaluation settings in HippoRAG \cite{jimenez2024hipporag} and HippoRAG2 \cite{gutierrez2025rag}, we employ the token-level F1 metric defined in the MuSiQue \cite{trivedi-etal-2022-musique}.

\textbf{Implementations.} For the implementation of HELP, we maintain strict architectural parity with the settings described in HippoRAG2 \cite{gutierrez2025rag} to ensure the validity of our comparative analysis. Specifically, \textit{Llama-3.3-70B-Instruct} \cite{llama3modelcard} powers both the knowledge extraction (NER/OpenIE) and response generation phases, while \textit{NV-Embed-v2} \cite{lee2024nv} serves as the primary retriever to bridge structural knowledge with semantic embeddings. Our QA module uses the top-5 retrieved passages as context for an LLM to generate the final answer. This same combination of extractor and retriever is applied across all reproduced Graph-based RAG baselines to ensure a fair evaluation.

\subsection{QA performance}
Table~\ref{tab:results} shows the performance of HELP compared to competitive baselines. The hyperparameter configurations for HELP are consistently applied across all datasets as detailed in Appendix~\ref{appendix: hyperparameter setting}. Overall, HELP sets a new state-of-the-art across the evaluated benchmarks, achieving a consistent performance gain over Graph-based RAG methods.

\textbf{Overall Superiority.} Compared to the previous leading Graph-based RAG method HippoRAG2, HELP yields a relative improvement of 1.3\% in average F1 score. Furthermore, HELP significantly outperforms the strongest large-scale embedding model NV-Embed-v2, with a 7.0\% relative gain in average F1, demonstrating that even state-of-the-art 7B embedding models can be significantly enhanced through our structural knowledge integration. When compared to traditional dense retrievers such as GTR, our method exhibits a substantial performance leap of 21.3\%, underscoring the vital necessity of bridging structural knowledge with semantic retrieval for complex question answering.

\begin{figure}[t!] 
    \centering
    \includegraphics[width=1\linewidth]{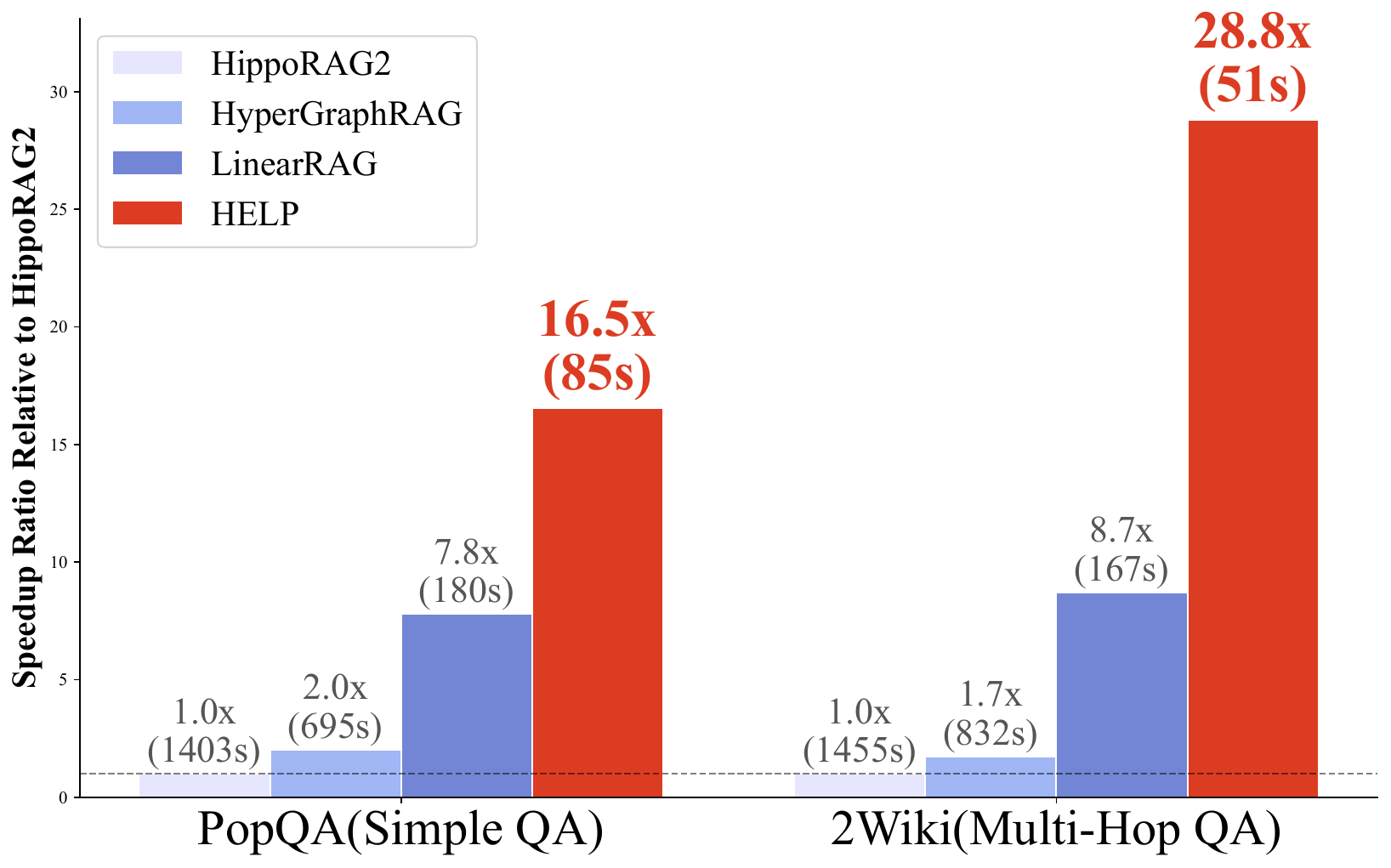} 
    \vspace{-18pt}
    \caption{Retrieval efficiency on PopQA (Simple QA) and 2Wiki (Multi-Hop QA). Absolute retrieval time (in seconds) for processing 1,000 queries are annotated above the bars, highlighting that HELP reduces the total retrieval latency to under 90 seconds.}
    \label{fig:speedup_comparison}
    \vspace{-10pt}
\vspace{-8pt}
\end{figure}

\textbf{Performance on Simple QA.} In single-hop scenarios represented by NQ and PopQA, HELP demonstrates superior retrieval precision. While the Hypernode Expansion mechanism is specifically engineered to navigate high-order relational dependencies in multi-hop reasoning, it notably maintains a competitive advantage in simpler, factoid-based tasks. Specifically, HELP achieves a 2.5\% relative improvement over the second-best performing method HippoRAG2 on PopQA dataset. Furthermore, it maintains a steady gain of 2.6\% to 3.4\% over the strongest 7B embedding models, such as NV-Embed-v2. This indicates that our knowledge-enhanced approach effectively anchors semantic search with precise evidence localization, ensuring that simple factoid queries benefit from the structural clarity of the knowledge graph while maintaining a decisive competitive edge over purely latent-space representations.

\textbf{Performance on Multi-Hop QA.} The advantages of HELP are most pronounced in complex reasoning tasks where relational dependencies are critical. On the 2Wiki dataset, HELP outperforms HippoRAG2 by 4.1\%. Compared to more recent methods like LinearRAG, HELP achieves a staggering 34.6\% relative improvement on the same benchmark, suggesting that our method more effectively navigates multi-hop paths than linear structures. Notably, while the indexing complexity of HyperGraphRAG rendered it unusable for the large-scale corpora of LV-Eval, HELP successfully processed the entire dataset, demonstrating both great reasoning capabilities and superior scalability.

\subsection{Retrieval Efficiency}

As illustrated in Fig.~\ref{fig:speedup_comparison}, we evaluated the retrieval time of HELP against three strong baselines: HippoRAG2, HyperGraphRAG, and LinearRAG.

Our method demonstrates overwhelming efficiency across both simple and multi-hop QA benchmarks. Specifically, on the PopQA dataset, HELP achieves a 16.5$\times$ speedup relative to HippoRAG2, drastically compressing the total retrieval time for 1,000 queries from 1,403s to a mere 85s. This performance gap widens even further on the complex 2Wiki dataset, where HELP attains a remarkable 28.8$\times$ speedup. Traditional graph-based baselines still suffer from the combinatorial explosion of graph traversals, requiring over 20 minutes to resolve dependencies, whereas HELP consistently completes the same tasks in under 90 seconds.

The significant efficiency gain of HELP stems from its HyperNode-based pruning strategy and a purely embedding-driven retrieval process. Unlike iterative graph methods that rely on LLMs for intermediate node filtering or path expansion, HELP bypasses these costly generative steps and the redundant neighbor explorations typical of traditional graph walks. Even when compared to LinearRAG, which was specifically designed to mitigate graph overhead, HELP further accelerates the process by approximately 2-3$\times$. This confirms that HELP successfully retains the structural reasoning benefits of knowledge graphs without succumbing to their traditional computational bottlenecks, proving its scalability for large-scale scenarios.

\begin{table}[t!]
\centering
\caption{Ablation study of the Hybrid Retrieval Strategy on the 2Wiki dataset. 
$M$ denotes the primary quota of passages assigned to the logical path-guided retrieval, while the total context size is fixed at 5. $M=0$ corresponds to a purely dense retrieval baseline.
}
\vspace{-5pt}
\label{tab:hybrid_ablation}
\small
\setlength{\tabcolsep}{12pt} 
\begin{tabular}{c c c c}
\toprule
\textbf{Quota(M)} & \textbf{EM(\%)} & \textbf{F1(\%)} &  \textbf{Recall@5(\%)} \\
\midrule
0 & 57.5 & 61.55 & 76.25 \\
1 & 57.7 & 62.19 & 76.52 \\
2 & 63.0 & 69.52 & 84.95 \\
3 & 64.3 & 71.01 & 89.00 \\
\rowcolor{blue!10} \textbf{4} & \textbf{66.6} & \textbf{73.90} & \textbf{92.15} \\
5 & 65.9 & 73.09 & 91.65 \\
\bottomrule
\end{tabular}
\vspace{-20pt}
\end{table}

\subsection{Ablation Study}

\textbf{1) Hybrid Retrieval Strategy. }\label{sec: hybrid strategy}
Table~\ref{tab:hybrid_ablation} underscores the synergy between structural precision and semantic coverage within our Hybrid Retrieval Strategy. To provide a multi-dimensional assessment, we report both Exact Match (EM), which measures the percentage of predicted answers that exactly match the ground truth, and Recall@5, which indicates whether at least one of the top-5 retrieved passages contains the gold evidence required to answer the query. The evaluation shows that performance peaks at $M=4$, a 20\%+ improvement over pure dense retrieval ($M=0$), proving that logical paths effectively resolve multi-hop dependencies. However, when semantic backfill is constrained ($M=5$), performance slightly declines. This non-monotonic trend suggests that a purely logical path-guided approach is vulnerable to graph incompleteness and structural noise; thus, maintaining a dense retrieval component is essential as a robust fallback to ensure comprehensive evidentiary support.
 
  


\begin{figure}[t]
    \centering
    \includegraphics[width=1.0\linewidth]{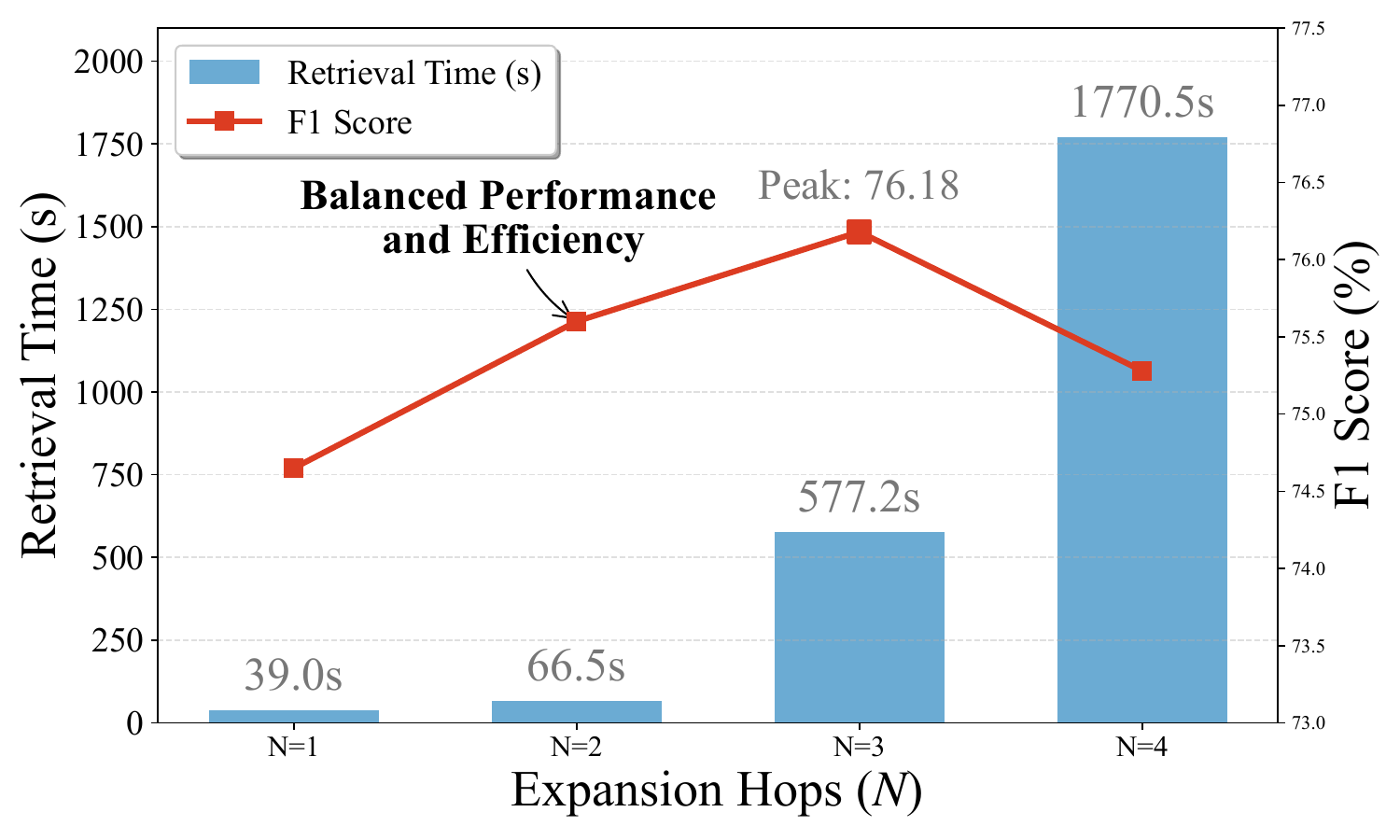}
    \vspace{-18pt}
    \caption{Experimental analysis of expansion hops $N$ on HotpotQA dataset. The bars indicate retrieval time (left axis), while the line tracks the QA F1 score (right axis). }
    \label{fig:efficiency_bar_line}
\vspace{-15pt}
\end{figure}

\textbf{2) Impact of Expansion Hops.}
To investigate the trade-off between retrieval latency and QA performance, we conducted an ablation study on the number of expansion hops, denoted as $N$. In Fig.~\ref{fig:efficiency_bar_line}, varying $N$ reveals a significant divergence between computational cost and QA performance.

On one hand, the retrieval time exhibits an exponential growth pattern. While the latency remains manageable at $N=1$ and $N=2$, it surges dramatically to $577.2$s at $N=3$ and becomes prohibitive at $N=4$. On the other hand, the F1 score follows a non-monotonic trajectory, yet remarkably remains above $74.5\%$ across all values of $N$, demonstrating HELP's robustness and strong performance. Specifically, performance peaks at $76.18\%$ when $N=3$, as deeper exploration retrieves more relevant context. However, increasing $N$ to $4$ leads to a performance degradation. This decline suggests that excessive expansion introduces significant noise and irrelevant information, which outweighs the benefits of broader context coverage. Given that the F1 score remains competitive even at lower hops, $N=2$ offers a compelling balance, achieving high accuracy with only a fraction of the computational overhead required for $N=3$.

\begin{figure}[t]
    \centering
    \includegraphics[width=0.9\linewidth]{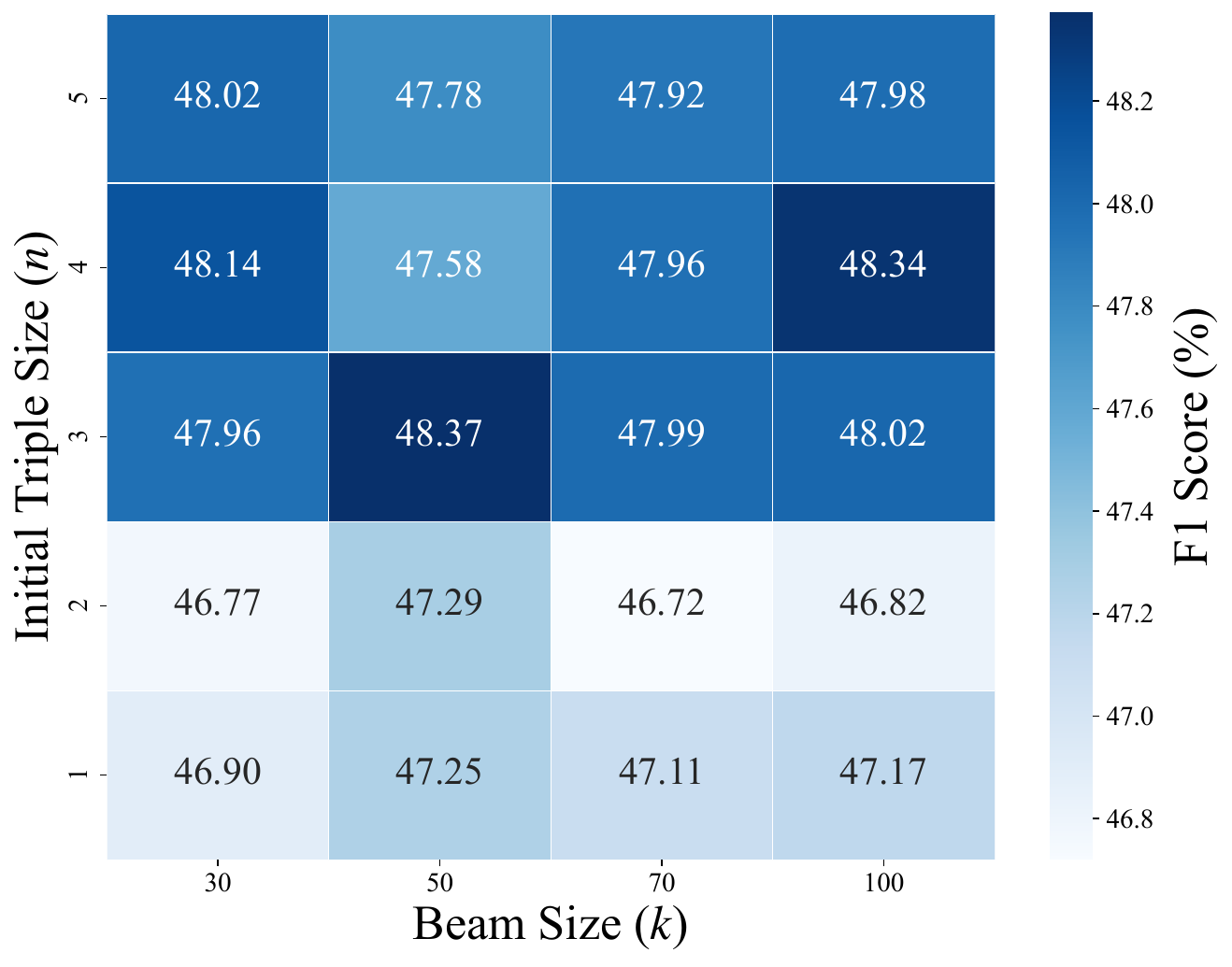}
    \vspace{-5pt}
    \caption{Hyperparameter sensitivity analysis of the Initial Triple Size ($n$) and Hypernode Beam Size ($k$) on MuSiQue dataset. The heatmap reports the F1 scores (\%), demonstrating the impact of varying the seed set size and pruning threshold.}
    \label{fig:hyperparam_sensitivity}
    \vspace{-18pt}
\end{figure}

\textbf{3) Hyperparameter Sensitivity Analysis.}
To investigate the impact of the search space configuration, we performed a grid search over the Initial Triple Size $n \in \{1, \dots, 5\}$ and the Hypernode Beam Size $k \in \{30, 50, 70, 100\}$. In Fig.~\ref{fig:hyperparam_sensitivity}, the model exhibits significant robustness to hyperparameter variations, with F1 scores fluctuating within a narrow range of 46.7\% to 48.4\%. This stability suggests that HELP is not overly sensitive to the specific choice of seed set size or pruning thresholds. Such minimal numerical variance, coupled with the consistent heatmap pattern, confirms that our method operates reliably without the need for extensive hyperparameter tuning, proving its practical scalability.

\textbf{4) Robustness Across Different Backbones.} 


To evaluate cross-model generalization, we tested HELP using \textit{Qwen3-30B-A3B-Instruct-2507}~\cite{qwen3technicalreport} as the backbone. The results, summarized in Table~\ref{tab:results_qwen}, demonstrate that HELP consistently outperforms recent graph-based RAG methods across both simple and complex reasoning tasks, achieving an average EM of 42.4\% and F1 of 52.6\%. Notably, HELP significantly outperforms HippoRAG2 in the long-context LV-Eval task, demonstrating superior evidentiary localization. These results confirm that HELP’s gains are model-agnostic and stem from its robust methodology rather than specific LLM capabilities, proving its versatility for diverse real-world deployments.

\vspace{-10pt}

\section{Conclusion}
In this paper, we introduced HELP, a robust GraphRAG framework that synergizes structural reasoning with retrieval efficiency. Through HyperNode Expansion and Logical Path-Guided Evidence Localization, HELP successfully transforms isolated knowledge triplets into integrated reasoning chains without incurring the high latency of exhaustive graph searches. Empirical results confirm that HELP significantly outperforms existing graph-based and dense retrieval methods in both simple and multi-hop reasoning accuracy and retrieval speed (up to 28.8$\times$ faster). Our work bridges the gap between structured knowledge representation and practical scalability, establishing a new standard for accurate and efficient GraphRAG methods.






\section*{Impact Statement}
This paper presents work on Retrieval-Augmented Generation (RAG) to advance the field of precise information retrieval for large language models. While our work may have various societal implications, we do not identify any concerns that warrant specific emphasis beyond those generally associated with large language models and information retrieval.


\bibliography{main}

@article{huang2025survey,
  title={A survey on hallucination in large language models: Principles, taxonomy, challenges, and open questions},
  author={Huang, Lei and Yu, Weijiang and Ma, Weitao and Zhong, Weihong and Feng, Zhangyin and Wang, Haotian and Chen, Qianglong and Peng, Weihua and Feng, Xiaocheng and Qin, Bing and others},
  journal={ACM Transactions on Information Systems},
  volume={43},
  number={2},
  pages={1--55},
  year={2025},
  publisher={ACM New York, NY}
}

@article{gao2023retrieval,
  title={Retrieval-augmented generation for large language models: A survey},
  author={Gao, Yunfan and Xiong, Yun and Gao, Xinyu and Jia, Kangxiang and Pan, Jinliu and Bi, Yuxi and Dai, Yixin and Sun, Jiawei and Wang, Haofen and Wang, Haofen},
  journal={arXiv preprint arXiv:2312.10997},
  volume={2},
  number={1},
  year={2023}
}

@article{lewis2020retrieval,
  title={Retrieval-augmented generation for knowledge-intensive nlp tasks},
  author={Lewis, Patrick and Perez, Ethan and Piktus, Aleksandra and Petroni, Fabio and Karpukhin, Vladimir and Goyal, Naman and K{\"u}ttler, Heinrich and Lewis, Mike and Yih, Wen-tau and Rockt{\"a}schel, Tim and others},
  journal={NeurIPS},
  volume={33},
  pages={9459--9474},
  year={2020}
}

@article{han2024retrieval,
  title={Retrieval-augmented generation with graphs (graphrag)},
  author={Han, Haoyu and Wang, Yu and Shomer, Harry and Guo, Kai and Ding, Jiayuan and Lei, Yongjia and Halappanavar, Mahantesh and Rossi, Ryan A and Mukherjee, Subhabrata and Tang, Xianfeng and others},
  journal={arXiv preprint arXiv:2501.00309},
  year={2024}
}

@article{peng2024graph,
  title={Graph retrieval-augmented generation: A survey},
  author={Peng, Boci and Zhu, Yun and Liu, Yongchao and Bo, Xiaohe and Shi, Haizhou and Hong, Chuntao and Zhang, Yan and Tang, Siliang},
  journal={ACM Transactions on Information Systems},
  year={2024},
  publisher={ACM New York, NY}
}

@article{plaat2025multi,
  title={Multi-step reasoning with large language models, a survey},
  author={Plaat, Aske and Wong, Annie and Verberne, Suzan and Broekens, Joost and Van Stein, Niki},
  journal={ACM Computing Surveys},
  year={2025},
  publisher={ACM New York, NY}
}

@article{zhang2025survey,
  title={A survey of graph retrieval-augmented generation for customized large language models},
  author={Zhang, Qinggang and Chen, Shengyuan and Bei, Yuanchen and Yuan, Zheng and Zhou, Huachi and Hong, Zijin and Chen, Hao and Xiao, Yilin and Zhou, Chuang and Dong, Junnan and others},
  journal={arXiv preprint arXiv:2501.13958},
  year={2025}
}

@article{han2025rag,
  title={Rag vs. graphrag: A systematic evaluation and key insights},
  author={Han, Haoyu and Ma, Li and Shomer, Harry and Wang, Yu and Lei, Yongjia and Guo, Kai and Hua, Zhigang and Long, Bo and Liu, Hui and Aggarwal, Charu C and others},
  journal={arXiv preprint arXiv:2502.11371},
  year={2025}
}

@article{xiang2025use,
  title={When to use graphs in rag: A comprehensive analysis for graph retrieval-augmented generation},
  author={Xiang, Zhishang and Wu, Chuanjie and Zhang, Qinggang and Chen, Shengyuan and Hong, Zijin and Huang, Xiao and Su, Jinsong},
  journal={arXiv preprint arXiv:2506.05690},
  year={2025}
}

@article{edge2024local,
  title={From local to global: A graph rag approach to query-focused summarization},
  author={Edge, Darren and Trinh, Ha and Cheng, Newman and Bradley, Joshua and Chao, Alex and Mody, Apurva and Truitt, Steven and Metropolitansky, Dasha and Ness, Robert Osazuwa and Larson, Jonathan},
  journal={arXiv preprint arXiv:2404.16130},
  year={2024}
}

@article{jimenez2024hipporag,
  title={Hipporag: Neurobiologically inspired long-term memory for large language models},
  author={Jimenez Gutierrez, Bernal and Shu, Yiheng and Gu, Yu and Yasunaga, Michihiro and Su, Yu},
  journal={Advances in Neural Information Processing Systems},
  volume={37},
  pages={59532--59569},
  year={2024}
}

@article{gutierrez2025rag,
  title={From rag to memory: Non-parametric continual learning for large language models},
  author={Guti{\'e}rrez, Bernal Jim{\'e}nez and Shu, Yiheng and Qi, Weijian and Zhou, Sizhe and Su, Yu},
  journal={arXiv preprint arXiv:2502.14802},
  year={2025}
}

@article{zhuang2025linearrag,
  title={LinearRAG: Linear Graph Retrieval Augmented Generation on Large-scale Corpora},
  author={Zhuang, Luyao and Chen, Shengyuan and Xiao, Yilin and Zhou, Huachi and Zhang, Yujing and Chen, Hao and Zhang, Qinggang and Huang, Xiao},
  journal={arXiv preprint arXiv:2510.10114},
  year={2025}
}

@article{sun2023think,
  title={Think-on-graph: Deep and responsible reasoning of large language model on knowledge graph},
  author={Sun, Jiashuo and Xu, Chengjin and Tang, Lumingyuan and Wang, Saizhuo and Lin, Chen and Gong, Yeyun and Ni, Lionel M and Shum, Heung-Yeung and Guo, Jian},
  journal={arXiv preprint arXiv:2307.07697},
  year={2023}
}

@article{ma2024think,
  title={Think-on-graph 2.0: Deep and interpretable large language model reasoning with knowledge graph-guided retrieval},
  author={Ma, Shengjie and Xu, Chengjin and Jiang, Xuhui and Li, Muzhi and Qu, Huaren and Guo, Jian},
  journal={arXiv e-prints},
  pages={arXiv--2407},
  year={2024}
}

@article{yang2024efficient,
  title={Efficient algorithms for personalized pagerank computation: A survey},
  author={Yang, Mingji and Wang, Hanzhi and Wei, Zhewei and Wang, Sibo and Wen, Ji-Rong},
  journal={IEEE Transactions on Knowledge and Data Engineering},
  volume={36},
  number={9},
  pages={4582--4602},
  year={2024},
  publisher={IEEE}
}

@inproceedings{bhagdev2008hybrid,
  title={Hybrid search: Effectively combining keywords and semantic searches},
  author={Bhagdev, Ravish and Chapman, Sam and Ciravegna, Fabio and Lanfranchi, Vitaveska and Petrelli, Daniela},
  booktitle={European semantic web conference},
  pages={554--568},
  year={2008},
  organization={Springer}
}

@article{wang2025balancing,
  title={Balancing the Blend: An Experimental Analysis of Trade-offs in Hybrid Search},
  author={Wang, Mengzhao and Tan, Boyu and Gao, Yunjun and Jin, Hai and Zhang, Yingfeng and Ke, Xiangyu and Xu, Xiaoliang and Zhu, Yifan},
  journal={arXiv preprint arXiv:2508.01405},
  year={2025}
}

@article{trivedi-etal-2022-musique,
    title = "{M}u{S}i{Q}ue: Multihop Questions via Single-hop Question Composition",
    author = "Trivedi, Harsh  and
      Balasubramanian, Niranjan  and
      Khot, Tushar  and
      Sabharwal, Ashish",
    editor = "Roark, Brian  and
      Nenkova, Ani",
    journal = "Transactions of the Association for Computational Linguistics",
    volume = "10",
    year = "2022",
    address = "Cambridge, MA",
    publisher = "MIT Press",
    doi = "10.1162/tacl_a_00475",
    pages = "539--554"
}

@article{wang2024rear,
  title={Rear: A relevance-aware retrieval-augmented framework for open-domain question answering},
  author={Wang, Yuhao and Ren, Ruiyang and Li, Junyi and Zhao, Wayne Xin and Liu, Jing and Wen, Ji-Rong},
  journal={arXiv preprint arXiv:2402.17497},
  year={2024}
}

@inproceedings{mallen-etal-2023-trust,
    title = "When Not to Trust Language Models: Investigating Effectiveness of Parametric and Non-Parametric Memories",
    author = "Mallen, Alex  and
      Asai, Akari  and
      Zhong, Victor  and
      Das, Rajarshi  and
      Khashabi, Daniel  and
      Hajishirzi, Hannaneh",
    editor = "Rogers, Anna  and
      Boyd-Graber, Jordan  and
      Okazaki, Naoaki",
    booktitle = "Proceedings of the 61st Annual Meeting of the Association for Computational Linguistics (Volume 1: Long Papers)",
    month = jul,
    year = "2023",
    address = "Toronto, Canada",
    publisher = "Association for Computational Linguistics",
    doi = "10.18653/v1/2023.acl-long.546",
    pages = "9802--9822"
}

@article{ho2020constructing,
  title={Constructing a multi-hop qa dataset for comprehensive evaluation of reasoning steps},
  author={Ho, Xanh and Nguyen, Anh-Khoa Duong and Sugawara, Saku and Aizawa, Akiko},
  journal={arXiv preprint arXiv:2011.01060},
  year={2020}
}

@inproceedings{yang2018hotpotqa,
  title={HotpotQA: A dataset for diverse, explainable multi-hop question answering},
  author={Yang, Zhilin and Qi, Peng and Zhang, Saizheng and Bengio, Yoshua and Cohen, William and Salakhutdinov, Ruslan and Manning, Christopher D},
  booktitle={EMNLP},
  pages={2369--2380},
  year={2018}
}

@article{yuan2024lv,
  title={Lv-eval: A balanced long-context benchmark with 5 length levels up to 256k},
  author={Yuan, Tao and Ning, Xuefei and Zhou, Dong and Yang, Zhijie and Li, Shiyao and Zhuang, Minghui and Tan, Zheyue and Yao, Zhuyu and Lin, Dahua and Li, Boxun and others},
  journal={arXiv preprint arXiv:2402.05136},
  year={2024}
}

@inproceedings{robertson1994some,
  title={Some simple effective approximations to the 2-poisson model for probabilistic weighted retrieval},
  author={Robertson, Stephen E and Walker, Steve},
  booktitle={SIGIR’94: Proceedings of the Seventeenth Annual International ACM-SIGIR Conference on Research and Development in Information Retrieval, organised by Dublin City University},
  pages={232--241},
  year={1994},
  organization={Springer}
}

@article{izacard2021unsupervised,
  title={Unsupervised dense information retrieval with contrastive learning},
  author={Izacard, Gautier and Caron, Mathilde and Hosseini, Lucas and Riedel, Sebastian and Bojanowski, Piotr and Joulin, Armand and Grave, Edouard},
  journal={arXiv preprint arXiv:2112.09118},
  year={2021}
}

@inproceedings{ni2022large,
  title={Large dual encoders are generalizable retrievers},
  author={Ni, Jianmo and Qu, Chen and Lu, Jing and Dai, Zhuyun and Abrego, Gustavo Hernandez and Ma, Ji and Zhao, Vincent and Luan, Yi and Hall, Keith and Chang, Ming-Wei and others},
  booktitle={EMNLP},
  pages={9844--9855},
  year={2022}
}

@article{thakur2021beir,
  title={Beir: A heterogenous benchmark for zero-shot evaluation of information retrieval models},
  author={Thakur, Nandan and Reimers, Nils and R{\"u}ckl{\'e}, Andreas and Srivastava, Abhishek and Gurevych, Iryna},
  journal={arXiv preprint arXiv:2104.08663},
  year={2021}
}

@article{li2023towards,
  title={Towards general text embeddings with multi-stage contrastive learning},
  author={Li, Zehan and Zhang, Xin and Zhang, Yanzhao and Long, Dingkun and Xie, Pengjun and Zhang, Meishan},
  journal={arXiv preprint arXiv:2308.03281},
  year={2023}
}

@inproceedings{muennighoff2024generative,
  title={Generative representational instruction tuning},
  author={Muennighoff, Niklas and Hongjin, SU and Wang, Liang and Yang, Nan and Wei, Furu and Yu, Tao and Singh, Amanpreet and Kiela, Douwe},
  booktitle={ICLR},
  year={2024}
}

@article{lee2024nv,
  title={Nv-embed: Improved techniques for training llms as generalist embedding models},
  author={Lee, Chankyu and Roy, Rajarshi and Xu, Mengyao and Raiman, Jonathan and Shoeybi, Mohammad and Catanzaro, Bryan and Ping, Wei},
  journal={arXiv preprint arXiv:2405.17428},
  year={2024}
}

@inproceedings{sarthi2024raptor,
  title={Raptor: Recursive abstractive processing for tree-organized retrieval},
  author={Sarthi, Parth and Abdullah, Salman and Tuli, Aditi and Khanna, Shubh and Goldie, Anna and Manning, Christopher D},
  booktitle={ICLR},
  year={2024}
}

@article{guo2024lightrag,
  title={Lightrag: Simple and fast retrieval-augmented generation},
  author={Guo, Zirui and Xia, Lianghao and Yu, Yanhua and Ao, Tu and Huang, Chao},
  journal={arXiv preprint arXiv:2410.05779},
  year={2024}
}

@article{luo2025hypergraphrag,
  title={HyperGraphRAG: Retrieval-Augmented Generation via Hypergraph-Structured Knowledge Representation},
  author={Luo, Haoran and Chen, Guanting and Zheng, Yandan and Wu, Xiaobao and Guo, Yikai and Lin, Qika and Feng, Yu and Kuang, Zemin and Song, Meina and Zhu, Yifan and others},
  journal={arXiv preprint arXiv:2503.21322},
  year={2025}
}

@article{llama3modelcard,
  title={Llama 3 Model Card},
  author={AI@Meta},
  year={2024},
  url = {https://github.com/meta-llama/llama3/blob/main/MODEL_CARD.md}
}

@inproceedings{trivedi2023interleaving,
  title={Interleaving retrieval with chain-of-thought reasoning for knowledge-intensive multi-step questions},
  author={Trivedi, Harsh and Balasubramanian, Niranjan and Khot, Tushar and Sabharwal, Ashish},
  booktitle={Proceedings of the 61st annual meeting of the association for computational linguistics (volume 1: long papers)},
  pages={10014--10037},
  year={2023}
}

@misc{hu2025graggraphretrievalaugmentedgeneration,
      title={GRAG: Graph Retrieval-Augmented Generation}, 
      author={Yuntong Hu and Zhihan Lei and Zheng Zhang and Bo Pan and Chen Ling and Liang Zhao},
      year={2025},
      eprint={2405.16506},
      archivePrefix={arXiv},
      primaryClass={cs.LG},
      url={https://arxiv.org/abs/2405.16506}, 
}

@misc{mavromatis2024gnnraggraphneuralretrieval,
      title={GNN-RAG: Graph Neural Retrieval for Large Language Model Reasoning}, 
      author={Costas Mavromatis and George Karypis},
      year={2024},
      eprint={2405.20139},
      archivePrefix={arXiv},
      primaryClass={cs.CL},
      url={https://arxiv.org/abs/2405.20139}, 
}

@article{yasunaga2021qa,
  title={QA-GNN: Reasoning with language models and knowledge graphs for question answering},
  author={Yasunaga, Michihiro and Ren, Hongyu and Bosselut, Antoine and Liang, Percy and Leskovec, Jure},
  journal={arXiv preprint arXiv:2104.06378},
  year={2021}
}

@article{jin2024graph,
  title={Graph chain-of-thought: Augmenting large language models by reasoning on graphs},
  author={Jin, Bowen and Xie, Chulin and Zhang, Jiawei and Roy, Kashob Kumar and Zhang, Yu and Li, Zheng and Li, Ruirui and Tang, Xianfeng and Wang, Suhang and Meng, Yu and others},
  journal={arXiv preprint arXiv:2404.07103},
  year={2024}
}

@inproceedings{jiang2025kg,
  title={Kg-agent: An efficient autonomous agent framework for complex reasoning over knowledge graph},
  author={Jiang, Jinhao and Zhou, Kun and Zhao, Wayne Xin and Song, Yang and Zhu, Chen and Zhu, Hengshu and Wen, Ji-Rong},
  booktitle={Proceedings of the 63rd Annual Meeting of the Association for Computational Linguistics (Volume 1: Long Papers)},
  pages={9505--9523},
  year={2025}
}

@article{zhao20252graphrag,
  title={E\^{} 2GraphRAG: Streamlining Graph-based RAG for High Efficiency and Effectiveness},
  author={Zhao, Yibo and Zhu, Jiapeng and Guo, Ye and He, Kangkang and Li, Xiang},
  journal={arXiv preprint arXiv:2505.24226},
  year={2025}
}

@misc{qwen3technicalreport,
      title={Qwen3 Technical Report}, 
      author={Qwen Team},
      year={2025},
      eprint={2505.09388},
      archivePrefix={arXiv},
      primaryClass={cs.CL},
      url={https://arxiv.org/abs/2505.09388}, 
}
\bibliographystyle{icml2026}

\newpage
\appendix
\onecolumn
\section{Appendix}

This appendix provides supplementary information to the main paper. The contents are organized as follows:
\begin{itemize}[nosep]
    \item Section~\ref{appendix: hyperparameter setting} details the core hyperparameter setting used in the HELP framework.
    \item Section~\ref{appendix: different backbones} presents additional experimental results across various retrieval baselines using the \textit{Qwen3-30B-A3B-Instruct-2507} backbone.
    \item Section~\ref{appendix: baselines} describes the implementation details and reproduction alignment of the baseline methods to ensure a fair comparison.    \item Section~\ref{appendix: case study} provides a qualitative analysis via a detailed case study on multi-hop reasoning performance.
\end{itemize}

\subsection{HyperParameter Setting}\label{appendix: hyperparameter setting}
The core hyperparameters for HELP were configured as follows.
\begin{table}[h]
    \centering
    \caption{Hyperparameter settings for the HELP framework.}
    \label{tab:hyperparameters}
    \small
    \renewcommand{\arraystretch}{1.2}
    \begin{tabular}{clc}
        \toprule
        \textbf{Symbol} & \textbf{Description} & \textbf{Value} \\
        \midrule
        $N$ & Number of expansion hops & 2 \\
        $n$ & Initial seed triple size & 3 \\
        $k$ & Hypernode beam search size & 50 \\
        $M$ & Quota for logical path-guided passages & 4 \\
        $L$ & Total context size ($M$ + Dense passages) & 5 \\
        \bottomrule
    \end{tabular}
\end{table}

\subsection{Different LLM Backbones}\label{appendix: different backbones}

To further validate that the performance gains of HELP are architectural rather than dependent on a specific language model, we conducted extensive evaluations using \textit{Qwen3-30B-A3B-Instruct-2507} as the generation backbone. The results, summarized in Table~\ref{tab:results_qwen}, demonstrate that HELP consistently outperforms recent graph-based retrieval methods across both simple and complex reasoning tasks.



\begin{table*}[h]
    \centering
    \caption{Performance comparison (EM / F1 scores) of various retrieval methods across different tasks, with all models evaluated using \textit{Qwen3-30B-A3B-Instruct-2507} as the backbone. In each cell, the first value represents the Exact Match (EM) score, and the second represents the F1 score. The best results are in \textbf{bold}. The $^\dagger$ symbol indicates a failure in graph construction due to large corpus size.}
    \label{tab:results_qwen}
    \small
    \vspace{-5pt}
    \begin{tabular}{l cccccc c}
        \toprule
        & \multicolumn{2}{c}{Simple QA} & \multicolumn{4}{c}{Multi-Hop QA} & \\
        \cmidrule(lr){2-3} \cmidrule(lr){4-7}
        Retrieval & NQ & PopQA & MuSiQue & 2Wiki & HotpotQA & LV-Eval & Avg \\
        \midrule
        \multicolumn{8}{c}{\textit{Graph-based RAG Methods}} \\
        \midrule
        HippoRAG2 \cite{gutierrez2025rag} & 40.4 / 54.4 & 42.2 / 56.1 & \textbf{34.3} / \textbf{45.8} & 60.2 / 69.4 & 58.4 / 72.9 & 7.26 / 9.67 & 40.5 / 51.4 \\
        LinearRAG \cite{zhuang2025linearrag} & 31.6 / 43.8 & 6.6 / 21.9 & 18.7 / 28.1 & 42.4 / 53.2 & 39.6 / 52.4 & 6.5/ 10.7 & 24.2 / 35.0 \\
        HyperGraphRAG \cite{luo2025hypergraphrag} & 30.3 / 42.1 & 7.9 / 22.5 & 22.7 / 36.9 & 45.5 / 58.6 & 54.3 / 70.2 & N/A$^\dagger$ & - \\
        \midrule
        \rowcolor{blue!10} \textbf{HELP (Ours)} & \textbf{43.0} / \textbf{56.9} & \textbf{44.5} / \textbf{57.4} & 33.7 / 44.7 & \textbf{62.0} / \textbf{69.5} & \textbf{60.4} / \textbf{73.9} & \textbf{10.5} / \textbf{12.9} & \textbf{42.4} / \textbf{52.6} \\
        \bottomrule
    \end{tabular}
\end{table*}

\subsection{Baseline Implementation Details}\label{appendix: baselines}
To ensure a fair and comprehensive comparison, all baseline methods we reproduced were evaluated under a strictly controlled environment. Specifically, we aligned these baselines with HELP by using an identical corpus, evaluation datasets, and the same LLM backbones for answer generation. Beyond these necessary alignments, all other internal configurations were kept strictly consistent with their original official implementations as reported in their respective literature. This ensures that the observed performance gains of HELP are attributable to its structural and logical innovations rather than disparate parameter tuning or modified baseline settings.

\newpage
\subsection{Case Study}\label{appendix: case study}

For this case study, HELP is configured with the hyperparameters specified in Table~\ref{tab:hyperparameters} and utilizes \textit{Llama-3.3-70B-Instruct} as the underlying LLM backbone. Table~\ref{tab:case_study_analysis} presents a qualitative comparison of multi-hop reasoning between HELP and several baselines. The query requires a 2-hop link: Princess Elene of Georgia $\to$ Solomon II of Imereti $\to$ Prince Archil of Imereti.

LinearRAG and HyperGraphRAG fail due to semantic distractions, retrieving irrelevant information. While HippoRAG2 retrieves the correct passages, it fails to produce the final answer. In contrast, our method, HELP, anchors the reasoning process by expanding from the Initial Triple Seed HyperNodes. This strategy effectively filters out distractor entities and successfully reconstructs the reasoning path to identify the final answer.

\definecolor{cgreen}{RGB}{0, 150, 0}
\definecolor{cred}{RGB}{180, 0, 0}
\newcommand{\cmark}{\textcolor{cgreen}{\ding{51}}} 
\newcommand{\xmark}{\textcolor{cred}{\ding{55}}}   

\newlist{tabitemize}{itemize}{1}
\setlist[tabitemize]{label=\textbullet, nosep, leftmargin=1em, after=\vspace{-0.5\baselineskip}, before=\vspace{-0.5\baselineskip}}

\begin{table*}[htbp]
    \centering
    \small
    \renewcommand{\arraystretch}{1.3}
    \caption{A case study illustrating how different methods handle a 2-hop query requiring cross-passage inference. HELP successfully leverages seed HyperNodes to anchor the correct reasoning path, whereas other baselines suffer from semantic drift or hallucination.}
    \begin{tabularx}{\textwidth}{
        >{\bfseries}p{0.18\textwidth} 
        >{\RaggedRight\arraybackslash}X 
    }
        \toprule
        Question & Who is the husband of Princess Elene Of Georgia? \\
        \midrule
        Ground Truth & Prince Archil of Imereti \\
        \midrule
        Reasoning Chain & $\text{Princess Elene of Georgia} \xrightarrow{\text{mother of}} \text{Solomon II of Imereti} \xrightarrow{\text{had father}} \text{Prince Archil of Imereti}$ \\
        \bottomrule
        \addlinespace[0.5em]

        
        LinearRAG & 
        \textbf{Retrieved Passages: } \newline
        1) \xmark \ Grand Duchess Elena Vladimirovna of Russia: ...husband was Prince Nicholas... \newline
        2) \xmark \ Prince Adarnase of Kartli: ...natural son of Levan of Kartli by a concubine... \newline
        3) \xmark \ Princess Charlotte of Württemberg: ...the wife of Grand Duke Michael Pavlovich of Russia... \newline
        4) \xmark \ Clous van Mechelen: ...is a Dutch musician, arranger, and actor... \newline
        5) \xmark \ Bernhard III, Prince of Anhalt-Bernburg: ...was the eldest son of bernhard ii... \newline
        \textbf{Prediction:} \xmark \ There is no direct information provided about the husband of Princess Elene of Georgia in the given passages. \newline \\
        \midrule

        HyperGraphRAG & 
        \textbf{Retrieved Passages: } \newline
        1) \xmark \ Gundobad: ...He was the husband of Caretene. \newline
        2) \xmark \ Eunoë (wife of Bogudes): Eunoë was the wife of Bogudes, King of Mauretania. \newline
        3) \xmark \ Engelbert III of the Mark: ...Adolph was the eldest son of Count Adolph II... \newline
        4) \xmark \ Megingoz of Guelders: ...He married Gerberga of Lorraine... \newline
        5) \xmark \ Princess Rodam of Kartli: ...Princess Rodam married King George VII... \newline
        \textbf{Prediction:} \xmark \ King George VII of Imereti \newline \\
        \midrule

        HippoRAG2 & 
        \textbf{Retrieved Passages: } \newline
        1) \cmark \ Princess Elene of Georgia: ...She was the mother of Solomon II of Imereti... \newline
        2) \xmark \ Grand Duchess Elena Vladimirovna of Russia: ...husband was Prince Nicholas... \newline
        3) \xmark \ Princess Rodam of Kartli: ...Princess Rodam married King George VII... \newline
        4) \cmark \ Solomon II of Imereti: ...He was born as David to Prince Archil of Imereti... \newline
        5) \xmark \ Grand Duchess Elena Pavlovna of Russia: ...his second wife Sophie Dorothea of Württemberg... \newline
        \textbf{Prediction:} \xmark \ None mentioned. \newline \\
        \midrule
        
        HELP & 
        \textbf{Expansion from Initial Triples as Seed HyperNodes} \newline
        ("princess elene of georgia", "daughter of", "heraclius ii") \newline 
        \textcolor{mygreen}{\textbf{("princess elene of georgia", "mother of", "solomon ii of imereti")}} \newline
        ("princess elene of georgia", "born in", "georgia") \newline 
        
        \textbf{Retrieved Passages: } \newline
        1) \cmark \  Princess Elene of Georgia: ...She was the mother of Solomon II of Imereti... \newline
        2) \cmark \  Solomon II of Imereti: ...He was born as David to Prince Archil of Imereti... \newline
        3) \xmark \  Grand Duchess Elena Vladimirovna of Russia: ...husband was Prince Nicholas... \newline
        4) \xmark \  Princess Rodam of Kartli: ...Princess Rodam married King George VII... \newline
        5) \xmark \  Grand Duchess Elena Pavlovna of Russia: ...his second wife Sophie Dorothea of Württemberg... \newline
        
        \textbf{Prediction:} \cmark \ Prince Archil of Imereti \newline \\
        \bottomrule
    \end{tabularx}
    
    \label{tab:case_study_analysis}
\end{table*}

\end{document}